\documentclass[11pt]{article}
\usepackage[T1]{fontenc}
\usepackage[utf8]{inputenc}
\usepackage{lmodern}
\usepackage{microtype}
\usepackage[a4paper,margin=25mm]{geometry}
\usepackage{amsmath,amssymb}
\usepackage{graphicx}
\usepackage{booktabs,tabularx,array}
\usepackage{xcolor}
\usepackage[numbers,sort&compress]{natbib}
\usepackage{caption}
\usepackage{enumitem}
\usepackage{tikz}
\usetikzlibrary{arrows.meta,positioning}
\usepackage{float}
\definecolor{inkblue}{HTML}{193B5A}
\definecolor{tealblue}{HTML}{187B83}
\definecolor{draftblue}{HTML}{245D8B}
\usepackage[colorlinks=true,linkcolor=inkblue,citecolor=tealblue,urlcolor=tealblue]{hyperref}
\usepackage[capitalize,noabbrev]{cleveref}
\hypersetup{pdftitle={MagNN-Retina preliminary manuscript with real public-data benchmarking},pdfauthor={Author information pending}}
\setlist{nosep,leftmargin=*}

\newcommand{\ind}{\mathbf{1}}

\DeclareMathOperator{\sigmoid}{sigmoid}

\usepackage{amsthm,longtable,placeins}
\newtheorem{proposition}{Proposition}
\newtheorem{corollary}[proposition]{Corollary}
\hypersetup{pdftitle={Reciprocity Can Erase Direction: Structural Boundaries and Controlled Utility of Magnetic Graph Networks}}
\title{Reciprocity Can Erase Direction:\\Structural Boundaries and Controlled Utility of Magnetic Graph Networks}
\author{Dawei Han, Zicheng Wang, Yaoxing Ren, Yi Lv, Rongqian Wu, and Fumin Wang}
\date{}
\hypersetup{pdfauthor={Dawei Han, Zicheng Wang, Yaoxing Ren, Yi Lv, Rongqian Wu, Fumin Wang}}
\begin{document}
\begin{titlepage}
\centering
\vspace*{5mm}
{\LARGE\bfseries Reciprocity Can Erase Direction:\par}
\vspace{3mm}
{\Large Structural Boundaries and Controlled Utility of Magnetic Graph Networks\par}
\vspace{12mm}
{\large
Dawei Han, PhD\textsuperscript{1,2,5\ensuremath{\dagger}},
Zicheng Wang, BS\textsuperscript{6\ensuremath{\dagger}},\\[3mm]
Yaoxing Ren, PhD\textsuperscript{1,2\ensuremath{\dagger}},
Yi Lv, MD\textsuperscript{1,2,3,4},\\[3mm]
Rongqian Wu, MD\textsuperscript{1,2,3*}, and
Fumin Wang, PhD\textsuperscript{1,2,3,4*}\par}
\vspace{10mm}
\begin{minipage}{\textwidth}
\small\raggedright
\setlength{\parskip}{3mm}
\textsuperscript{1}National Local Joint Engineering Research Center for Precision Surgery \& Regenerative Medicine, The First Affiliated Hospital of Xi'an Jiaotong University, Xi'an, PR China\par
\textsuperscript{2}School of Future Technology, Xi'an Jiaotong University, Xi'an, PR China\par
\textsuperscript{3}Med-X Research Institute, The First Affiliated Hospital of Xi'an Jiaotong University, Xi'an, PR China\par
\textsuperscript{4}Department of Hepatobiliary Surgery, The First Affiliated Hospital of Xi'an Jiaotong University, Xi'an, PR China\par
\textsuperscript{5}Office of Global promotion, Division of International Cooperation and Exchanges, Xidian University\par
\textsuperscript{6}Arts and Science Department, Cornell University, New York 14850, USA\par
\vspace{4mm}
\textsuperscript{\ensuremath{\dagger}}These authors contributed equally.\par
\textsuperscript{*}Corresponding authors: Rongqian Wu and Fumin Wang.\\
Correspondence email: \href{mailto:fwang1991@xjtu.edu.cn}{fwang1991@xjtu.edu.cn}\par
\end{minipage}
\vfill
\end{titlepage}

\begin{abstract}
Magnetic graph networks encode direction through phase, but a learned graph can erase the intended directional mechanism before prediction. We characterize this failure for support-based magnetic Laplacians. Exact identities relate imaginary energy and distance from zero phase to amplitude-weighted reciprocity. Top-$k$ selection forces complete collapse when the observed-node count satisfies $m\leq k+1$, and dense support imposes a lower bound on reciprocal-arc fraction. We then construct paired propagation examples with identical node features and symmetric amplitudes but opposite labels, giving a provable chance-level limit for direction-erasing representations. In 72 fixed neural fits, support phase achieved test AUROC 1.0000 on a unilateral ring but 0.5000 on reciprocal ring and complete supports; continuous score phase and a real directed comparator achieved 1.0000 throughout. All models solved the direction-free control. These are oracle-direction mechanism tests, not evidence of learned clinical direction. In the preserved public DeepDRiD development study, continuous phase was active but its robustness difference from zero phase was only $+0.0036$ (descriptive 95\% interval $-0.0014$ to $+0.0095$) on validation patients used for checkpoint selection. The contribution is a structural and experimental framework for distinguishing erased direction from unestablished task utility. It provides a precise circumstance in which magnetic phase preserves useful information, without asserting universal or clinical superiority.
\end{abstract}

\section{Introduction}
Magnetic graph networks encode edge orientation through complex phases in a Hermitian operator \citep{zhang2021magnet}. This is attractive when directional relations are meaningful but an undirected representation loses them. In learned graphs, however, direction is not an intrinsic guarantee: a model first constructs scores, selects neighbors, and converts those selections into an operator. A failure at that conversion can make the intended phase mechanism unavailable before the predictor is optimized.

This paper examines a concrete failure mode. When phase is defined by the difference between binary edge supports, reciprocal selection sets phase to zero even if the two underlying scores differ. Complete top-$k$ support necessarily has this property. Missing observations can therefore turn an otherwise sparse construction into a complete graph without changing the requested $k$. A nominally magnetic architecture can then operate with exactly its zero-phase Laplacian.

\begin{quote}
\textbf{Core claim.} Reciprocal top-$k$ support can erase antisymmetric edge information from a support-based magnetic operator. Exact structural diagnostics identify when this occurs; paired directional tasks distinguish that representational failure from a lack of demonstrated benefit on a particular prediction task.
\end{quote}

\begin{figure}[htbp]
\centering
\includegraphics[width=\linewidth]{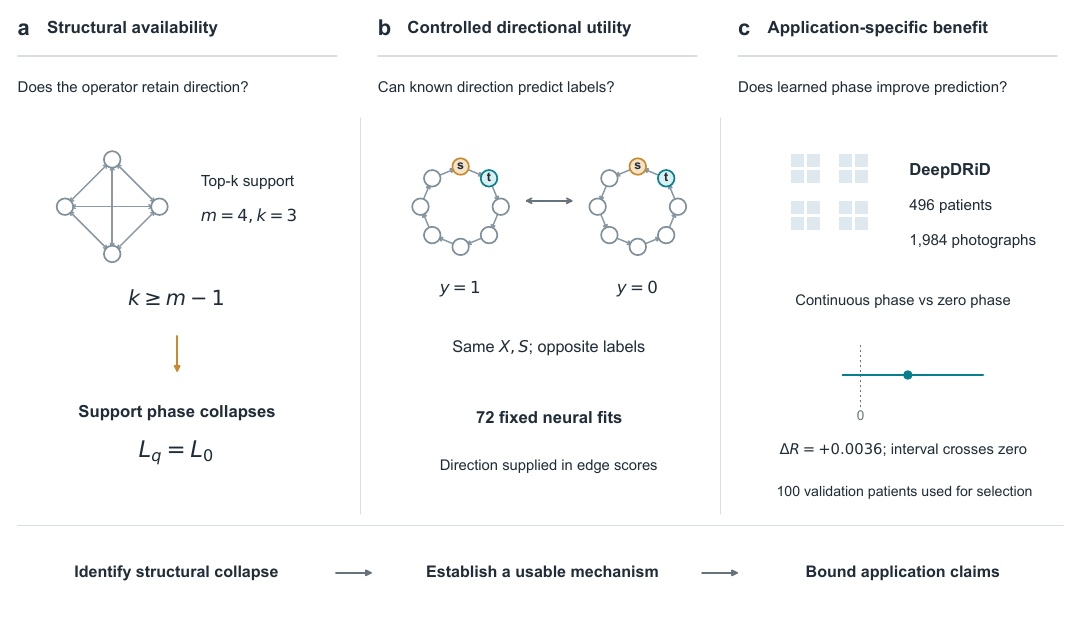}
\caption{Research roadmap. \textbf{a,} A complete four-node support illustrates the structural collapse boundary for support-based phase. \textbf{b,} Direction-reversed synthetic pairs share node features and symmetric amplitudes but have opposite propagation labels; $s$ denotes the source and $t$ the target. Direction is supplied in edge scores. \textbf{c,} The public DeepDRiD cohort tests application-specific utility. The small interval shows the continuous-minus-zero mean-readout contrast on 100 validation patients used for checkpoint selection: $\Delta R=+0.0036$, descriptive 95\% interval $[-0.0014,+0.0095]$. The squares depict the view/region organization, not patient photographs. The arrows summarize the argument, not a transfer of data between experiments.}
\label{fig:v08-roadmap}
\end{figure}

The study is organized around three distinct questions: structural availability, controlled directional utility, and application-specific benefit (\cref{fig:v08-roadmap}). Each stage supplies a different type of evidence.

We develop this claim through three connected contributions. First, we give exact phase-energy identities and a top-$k$/node-count boundary, separating support reciprocity from amplitude-weighted reciprocity. Second, we construct paired examples with identical node features and symmetric amplitudes but opposite propagation labels. These give a provable chance-level limit for the direction-erasing representation and an explicit magnetic decoder when orientation survives. A fixed neural experiment tests this mechanism, including a real-valued directed comparator. Third, we reinterpret an audited public retinal-image experiment as a test of task-dependent utility after phase availability has been checked. Its inconclusive comparative gains do not negate the synthetic mechanism and do not establish that images lack learnable direction.

Our claim concerns representation and experimental attribution, rather than universal accuracy superiority. The magnetic Laplacian and its directed-graph motivation are established \citep{zhang2021magnet}; incoming/outgoing real-valued aggregation is also a viable route to directional learning \citep{rossi2024direction}. We do not claim to invent either approach or to be the first to observe that symmetric support has zero phase. The contribution is the quantitative collapse analysis, its interaction with learned top-$k$ graphs and missing observations, and a controlled protocol that makes competing explanations of a null result distinguishable.

\section{Structural availability of magnetic phase}
\subsection{Operator and assumptions}
Consider $m$ observed nodes with no self edges. Let $B\in\{0,1\}^{m\times m}$ denote selected support and let $W_{ij}>0$ exactly when $B_{ij}=1$, with $W_{ij}=0$ otherwise. We use $W_{ij}$ for a message from $j$ to $i$. Scores may be learned and asymmetric; for example, $W_{ij}=B_{ij}\sigmoid(s_{ij})$. Define
\begin{equation}
 S=\tfrac12(W+W^\top),\quad d_i=\sum_j S_{ij},\quad
 A_{ij}=\frac{S_{ij}}{\sqrt{d_i d_j}},\quad
 L_q=\Pi-A\odot e^{\mathrm{i}2\pi q(B-B^\top)},
 \label{eq:v07-operator}
\end{equation}
where entries involving zero degrees are zero, $\Pi_{ii}=\ind(d_i>0)$, and exponentiation is entrywise. Strict positivity on selected edges is needed for the support-level equivalences below; under numerical underflow, the energy identities remain valid but a selected zero-weight edge need not contribute.

Let $U=\{(i,j):B_{ij}\ne B_{ji}\}$ be the ordered entries corresponding to unilateral connections. Define
\begin{equation}
 E_U=\sum_{(i,j)\in U}A_{ij}^2,\qquad E_A=\|A\|_F^2,
 \qquad r_E=1-\frac{E_U}{E_A}\quad(E_A>0).
 \label{eq:v07-energy}
\end{equation}
Thus $r_E$ weights reciprocity by normalized amplitude energy. It generally differs from the reciprocal-arc fraction $r_B=\sum_{ij}B_{ij}B_{ji}/\sum_{ij}B_{ij}$.

\begin{proposition}[Exact phase-energy identities]
For the operator in \cref{eq:v07-operator},
\begin{align}
 \|\Im L_q\|_F^2&=\sin^2(2\pi q)E_U,\label{eq:v07-imag}\\
 \|L_q-L_0\|_F^2&=4\sin^2(\pi q)E_U.\label{eq:v07-distance}
\end{align}
For $0<q<1/2$, $\Im L_q=0$ if and only if $B=B^\top$.
For $0<q\leq1/2$, $L_q=L_0$ if and only if $B=B^\top$.
When $E_A>0$, the normalized operator distance is exactly
\begin{equation}
 \frac{\|L_q-L_0\|_F}{\|A\|_F}=2|\sin(\pi q)|\sqrt{1-r_E}.
 \label{eq:v07-normalized}
\end{equation}
\label{prop:v07-energy}
\end{proposition}
\begin{proof}
Reciprocal pairs have zero phase and contribute nothing to either norm. On unilateral entries the phase is $\pm2\pi q$. Squaring the imaginary part yields $A_{ij}^2\sin^2(2\pi q)$; subtracting the zero-phase entry gives $A_{ij}^2|e^{\pm\mathrm{i}2\pi q}-1|^2=4A_{ij}^2\sin^2(\pi q)$. Summing proves the identities. Positive supported amplitudes imply $E_U=0$ exactly when there are no unilateral pairs. The specified charge ranges and division by $E_A$ give the remaining statements.
\end{proof}

The endpoint $q=1/2$ is consequential: an asymmetric support has zero imaginary part but can yield a real signed operator different from $L_0$. Consequently, an imaginary-norm diagnostic alone cannot detect every kind of phase effect. At $q=0$, equality to $L_0$ is trivial. We call $L_q=L_0$ \emph{operator collapse}; this definition avoids conflating zero imaginary part with complete phase removal.

\begin{proposition}[Top-$k$ structural boundary]
Suppose each node selects exactly $k_{\mathrm{eff}}=\min(k,m-1)$ distinct other observed nodes. If $k\geq m-1$, then support is complete and $L_q=L_0$ for every $q$. For $1\leq k_{\mathrm{eff}}\leq m-1$,
\begin{equation}
 r_B\geq\max\!\left\{0,\,2-\frac{m-1}{k_{\mathrm{eff}}}\right\}.
 \label{eq:v07-bound}
\end{equation}
\label{prop:v07-topk}
\end{proposition}
\begin{proof}
Complete support is reciprocal. Otherwise let $r$ be the number of reciprocal unordered pairs and $u$ the number of unilateral arcs. There are $mk_{\mathrm{eff}}=2r+u$ arcs, while $r+u\leq m(m-1)/2$. Therefore $u\leq m(m-1-k_{\mathrm{eff}})$ and $u\leq mk_{\mathrm{eff}}$. Substitution in $r_B=1-u/(mk_{\mathrm{eff}})$ proves the bound.
\end{proof}

At $k=0$ or $m\leq1$, the graph has no edges and its operator is zero. For $k<m-1$, nonzero phase is possible but not assured: a reciprocal ring with two neighbors already collapses for $m>3$. Moreover, \cref{eq:v07-bound} controls an unweighted count; it cannot be substituted for $r_E$ in \cref{eq:v07-normalized} without assumptions about amplitudes. High reciprocal counts can coexist with energy concentrated on a few unilateral edges.

\begin{corollary}[Missing observations can force collapse]
For a fixed requested $k$, any observed-node count $2\leq m\leq k+1$ forces complete reciprocal support. With four regional nodes per retained photograph and $k=3$, a single photograph lies exactly on this boundary.
\end{corollary}

These statements are pointwise in the constructed graph and thus hold at every learned parameter setting satisfying the assumptions. For a fixed network that uses direction only through $L_q$, replacing $q$ by zero on a collapsed graph leaves its prediction unchanged. This is not a claim that separately trained charged and zero-phase models must have identical parameters or predictions after training on other, phase-active graphs.

\begin{figure}[htbp]
\centering
\includegraphics[width=\linewidth]{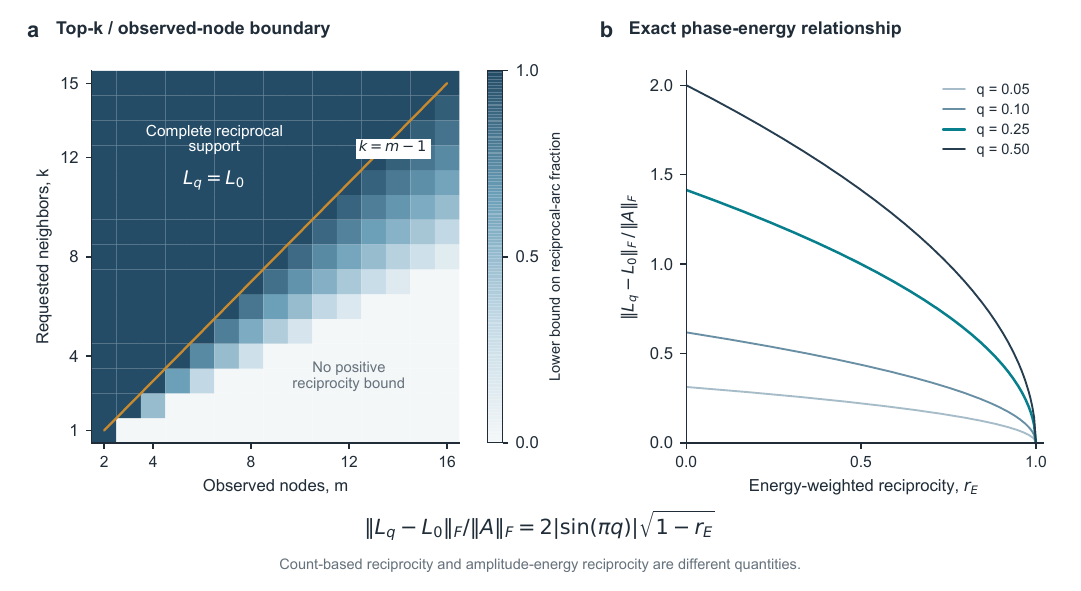}
\caption{Analytic structural boundaries. \textbf{a,} The reciprocal-arc fraction is bounded below by $\max\{0,2-(m-1)/k_{\mathrm{eff}}\}$, with $k_{\mathrm{eff}}=\min(k,m-1)$. Complete reciprocal support occurs on and above the orange boundary, where support phase collapses. The pale region indicates a zero lower bound, not guaranteed asymmetry. \textbf{b,} Exact normalized operator distance for four fixed charges. Its energy-weighted reciprocity $r_E$ differs from the count-based quantity in panel a. These are analytic values, not fitted experimental measurements.}
\label{fig:v07-boundary}
\end{figure}

\subsection{Continuous scores change the collapse condition}
A score-based phase uses
\begin{equation}
 \Theta_{ij}=2\pi q\tanh((s_{ij}-s_{ji})/\tau),\qquad
 L_\Theta=\Pi-A\odot e^{\mathrm{i}\Theta},\quad \tau>0.
 \label{eq:v07-continuous}
\end{equation}
Only the selected edge union matters, since $A$ is zero elsewhere. Antisymmetry of $\Theta$ preserves Hermitian structure. Directly,
\begin{equation}
 \|L_\Theta-L_0\|_F^2=4\sum_{ij} A_{ij}^2\sin^2(\Theta_{ij}/2).
 \label{eq:v07-cont-energy}
\end{equation}
Reciprocal support no longer suffices for collapse: unequal reciprocal scores can retain phase even on complete support. This is a change of representation, not a consequence of \cref{prop:v07-topk} for the same operator. The phase path is differentiable on retained edges; hard top-$k$ selection remains discrete. Nonzero phase or gradient alone does not establish useful prediction, causal direction, or a benefit over real directed aggregation.

\section{A paired test of directional utility}
\subsection{Identifiability before optimization}
We use a controlled task where direction has known predictive content. On an eight-node ring, mark a source and one of its two adjacent nodes as the target. A pulse follows a ground-truth directed cycle for one step. The binary label states whether it reaches the target. Each example is paired with the reverse cycle, using the same features, source, target, and node permutation; its label is reversed. The direction is supplied through edge scores, not inferred from identical node features.

Scores are 2 along the ground-truth cycle, 1 along the reverse edges, and $-2$ on other off-diagonal pairs. Diagonal entries are excluded. Row-wise top-$k$ selection followed by sigmoid weighting produces $W$. For $k=1$, only the cycle is selected; for $k=2$, both ring directions are selected; for $k=7$, support is complete. Each twin pair has the same $S$ for all three settings. At $k=2$ and 7, $W$ remains asymmetric although support is reciprocal. At these settings the ground-truth cycle is the higher-score direction; the weaker reverse edges are part of the observed weighted graph, not the ground-truth one-step propagation process.

\begin{proposition}[Paired representation gap]
On a balanced set of paired opposite-label examples with identical $(X,S)$, every deterministic predictor that depends only on $(X,L_0)$ produces equal scores within each pair. Its empirical AUROC and threshold accuracy are exactly $1/2$, and its mean binary cross-entropy is at least $\log2$. The same conclusion holds for support-phase predictors on the reciprocal $k=2$ and $k=7$ constructions. With $q=1/4$, the support-phase operator at $k=1$, and the continuous-phase operator at all three $k$, admit a one-step decoder with perfect accuracy on this construction.
\label{prop:v07-gap}
\end{proposition}
\begin{proof}
The direction-erasing representation assigns a common score to each pair. The positive and negative empirical score multisets are therefore identical, implying AUROC $1/2$, including half credit for ties. Any fixed threshold yields one correct label per pair. Pairwise cross-entropy is $-\tfrac12\log(p(1-p))\geq\log2$. Support collapse gives the same representation for reciprocal support. For the constructive part, let $e_s$ be the source pulse. The sign of $(-\Im L\, e_s)_t$ is positive when the target is the successor and negative for the reverse direction: the selected normalized amplitude is positive and the relevant phase lies strictly between $0$ and $\pi$. The target indicator locates $t$ independently of node order. Thresholding this sign recovers the label.
\end{proof}

The proposition describes a representational possibility, not a guarantee of gradient-based optimization. We therefore also fit small neural predictors. A direction-free control uses the sign of an observed target feature as the label; twins share that label. This control checks that collapse does not imply inability to learn any task. It is deliberately simple and is not evidence about the complexity of natural tasks.

\subsection{Locked neural protocol}
Each node has four features: source and target indicators and two noise channels. The target value of one noise channel has a sign that alternates deterministically with pair index, ensuring balance, and a random magnitude uniform on $[0.5,1.5)$; it determines only the direction-free label. Other noise values are standard normal. Node permutations are sampled independently across pairs. Pair IDs are used only for auditing and never as model inputs. There are 256 training, 64 validation, and 128 test pairs (512, 128, and 256 graphs) per task. Twin members remain within the same split, and split generators are independent. The two tasks share the graph/feature realizations and differ in labels, enabling a direct task control.

All four models use a width-16 linear/ReLU node encoder and a target-conditioned readout with a $32\rightarrow16\rightarrow1$ head. Magnetic models use one order-1 complex Chebyshev layer with a complex residual and separate real/imaginary ReLU; real and imaginary target embeddings are concatenated. The zero-phase control has the same complex parameters and architecture. The real directed model uses separate normalized incoming/outgoing messages and one residual layer, concatenating a zero imaginary block at readout. Parameter counts and their limits are reported with the results; matching nominal readout dimensions does not equalize effective capacity.

The four fixed operators are zero phase, support phase ($q=0.25$), continuous phase ($q=0.25$, $\tau=1$), and real directed aggregation. All receive the same scored graph before their declared conversion; no operator is given labels. Each task/operator/$k$ combination uses three initialization seeds, totaling 72 fits. Each fit runs 30 epochs of AdamW with learning rate 0.003, weight decay 0.0001, batch size 64, and gradient clipping at norm 5. The earliest minimum validation BCE selects the saved checkpoint. No charge search, test-directed tuning, or seed exclusion is performed. Test AUROC and accuracy are reported per fit and summarized by the mean and range across seeds, without treating seed variation as population uncertainty.

This is an \emph{oracle-direction positive control}. Because twins have identical $X$, a deterministic relation builder using $X$ alone could not recover their opposite orientations. Success with supplied scores demonstrates that an operator can retain and use a known directional signal; it does not demonstrate that a learned retinal graph recovers such a signal.

\section{Synthetic results: a structural failure and its remedy}
All 72 planned fits completed, yielding 18,432 test predictions across runs. Each task contains only 128 unique test pairs; repeated seeds and models do not multiply its sample size. All saved checkpoints reproduced their selected validation BCE and all saved test probabilities exactly under replay. An independent pairwise-ranking implementation verified all 72 AUROCs.

\begin{table}[htbp]
\centering
\caption{Synthetic test AUROC, averaged across all three initialization seeds. Top rows: direction-dependent propagation. Bottom rows: direction-free feature control. Test graphs are held out from training and checkpoint selection. All seed-level results and accuracies appear in the appendix.}
\label{tab:v07-synthetic}
\begin{tabular}{lrrrr}
\toprule
$k$ / task & Zero phase & Support phase & Continuous & Real directed \\
\midrule
1 & 0.5000 & 1.0000 & 1.0000 & 1.0000 \\
2 & 0.5000 & 0.5000 & 1.0000 & 1.0000 \\
7 & 0.5000 & 0.5000 & 1.0000 & 1.0000 \\
\midrule
1 (feature) & 1.0000 & 1.0000 & 1.0000 & 1.0000 \\
2 (feature) & 1.0000 & 1.0000 & 1.0000 & 1.0000 \\
7 (feature) & 1.0000 & 1.0000 & 1.0000 & 1.0000 \\
\bottomrule
\end{tabular}
\end{table}

On the directional task, support phase with $k=1$ achieved AUROC and accuracy of 1.0000 in every seed, while the matched zero-phase model achieved 0.5000. With reciprocal support at $k=2$ and 7, the support-phase model fell to chance, as predicted by \cref{prop:v07-gap}. Continuous phase achieved AUROC and accuracy of 1.0000 at all three settings. The real directed comparator also achieved 1.0000 throughout. Thus, the demonstrated advantage is relative to direction-erasing representations; these observations do not establish superiority over real directed learning.

All models achieved AUROC and accuracy of 1.0000 on the direction-free feature control, including collapsed support-phase and zero-phase models. Collapse therefore removes the ability to distinguish the designed orientations, rather than rendering the network generally untrainable.

Magnetic and zero-phase models each had 2,161 trainable parameters; the real directed comparator had 2,801. The exact chance-level argument is independent of parameter count for predictors restricted to the same erased representation. We do not infer computational efficiency or optimal capacity from these small implementations.

Two raw float32 runs (zero and support phase at $k=7$, seed 20260922) had AUROC 0.5000305 instead of exactly 0.5. Their twin inputs and Laplacians were bitwise identical, but a single twin pair differed by $5.96\times10^{-8}$ in predicted probability due to floating-point evaluation; accuracy remained 0.5. Averaging scores within pairs restored AUROC 0.5 as a diagnostic. All reported metrics use the original probabilities, and the complete table retains five decimal places. This numerical effect does not represent recovered directional information.

\begin{figure}[htbp]
\centering
\includegraphics[width=\linewidth]{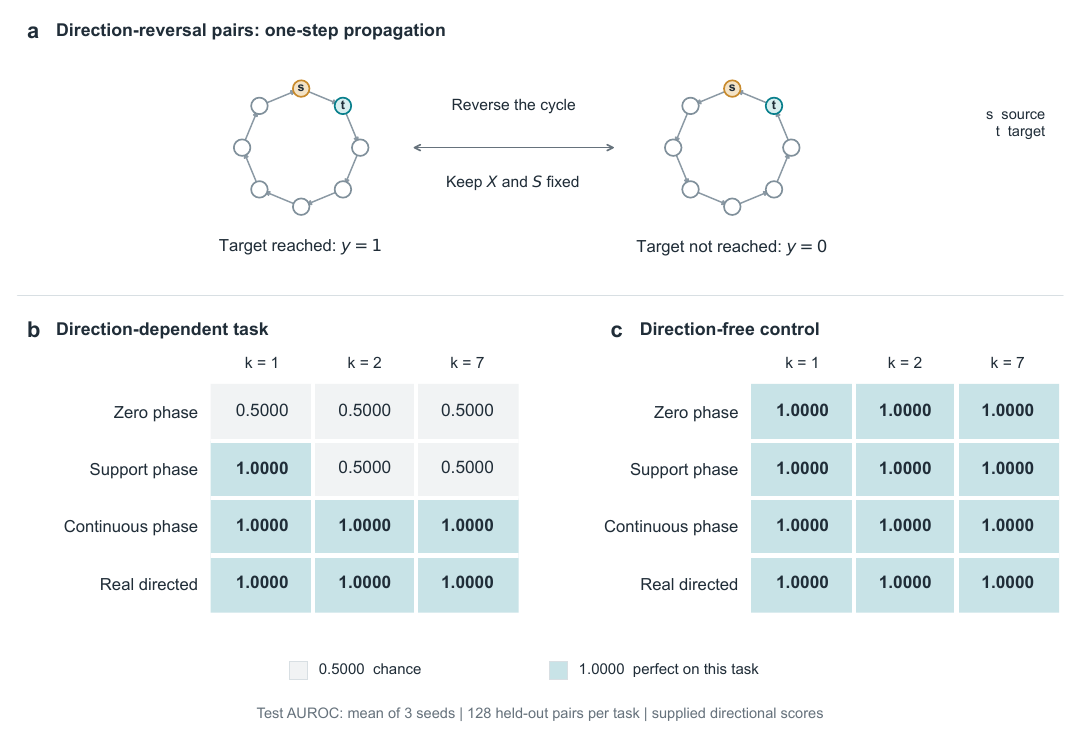}
\caption{Controlled directional utility. \textbf{a,} An illustrative source--target pair on an eight-node cycle. Reversal changes whether the source pulse reaches the adjacent target in one step, while preserving node features and symmetric amplitudes. \textbf{b,c,} Complete test AUROC matrices for the propagation task and direction-free feature control. Each cell is the mean of three initialization seeds on 128 held-out pairs; all 72 fits are retained. Columns correspond to unilateral ($k=1$), reciprocal-ring ($k=2$), and complete ($k=7$) support. Color encodes AUROC, not model identity. Values are rounded to four decimals; unrounded per-seed values, including the small float32 departures from 0.5, are retained in the source data and appendix. Directional scores are supplied inputs.}
\label{fig:v07-synthetic}
\end{figure}

\begin{figure}[htbp]
\centering
\includegraphics[width=\linewidth]{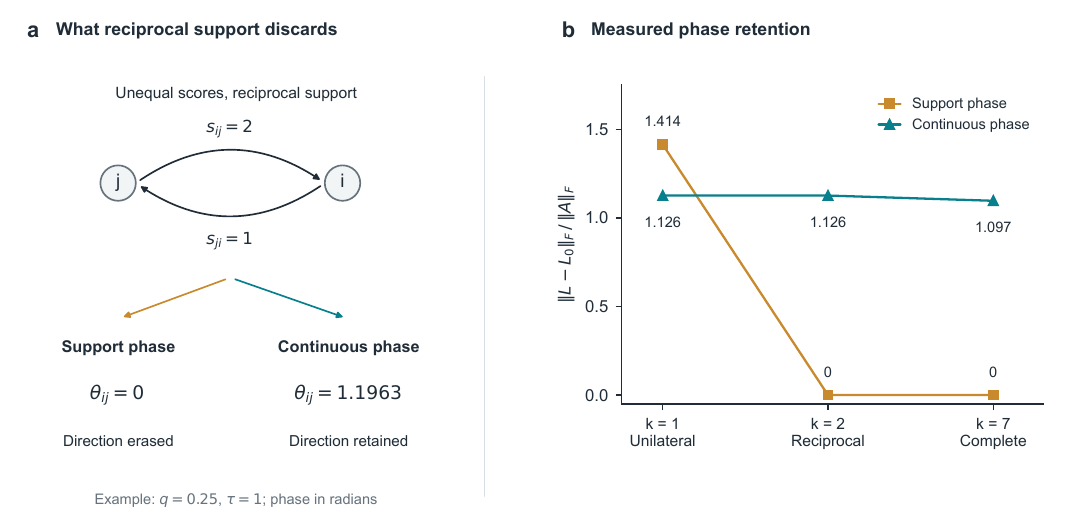}
\caption{How continuous phase preserves reciprocal score information. \textbf{a,} An illustrative pair has $s_{ij}=2$ and $s_{ji}=1$ with reciprocal binary support. At $q=0.25$ and $\tau=1$, support phase is zero whereas score-based phase is $\theta_{ij}=1.1963$ radians; the reverse entry has the opposite phase. \textbf{b,} Measured operator distance on the paired benchmark, normalized by $\|A\|_F$. Reciprocal support erases support phase at $k=2$ and 7, while continuous phase retains the asymmetry. Numeric labels are rounded; plotted values use the full recorded precision. Nonzero distance is a structural property, and the separate propagation experiment establishes utility on its designed task.}
\label{fig:v07-operators}
\end{figure}

Operator measurements agreed with the mechanism (\cref{fig:v07-operators}). Support reciprocity and energy reciprocity were both zero at $k=1$ and one at $k=2$ and 7. Normalized support-phase distance from $L_0$ was $\sqrt{2}$, 0, and 0, respectively. Continuous-phase distance was 1.1262, 1.1262, and 1.0966. The maximum discrepancy in the squared-norm identity was $3.55\times10^{-15}$. The explicit one-step sign decoder classified every held-out pair correctly for continuous phase at every $k$.

\FloatBarrier
\section{Real-data test: availability does not establish utility}
\subsection{Audited public cohort and development design}
The real-data component uses the public DeepDRiD repository \citep{liu2022deepdrid}, revision \texttt{56d8af7}, under CC BY-SA 4.0. Four training patients with ambiguous filename/eye correspondence were excluded before fitting, leaving 496 patients and 1,984 photographs: 296 training, 100 validation, and 100 test patients. The endpoint is patient-level maximum gradable-eye DR severity of at least grade 2; it does not include a diabetic macular edema criterion. There are four photographs per patient. No patient-level clinical covariates, omics, or follow-up outcomes are included.

Frozen ImageNet ResNet-18 features \citep{he2016resnet} provide four regional nodes per photograph, at most 16 per patient. A width-128 query/key relation builder selects $k=2$ neighbors. Magnetic models use two order-2 complex layers. A preceding $k=3$ diagnostic on single-view, four-node graphs observed reciprocal support, zero imaginary operator norm, and unchanged predictions when charge was removed. This is the exact boundary of \cref{prop:v07-topk}, rather than evidence that the endpoint lacks directional content. Reducing $k$ to 2 removed forced collapse in that diagnostic, but did not itself establish accuracy benefit.

The preserved development extension screened $q\in\{0,0.05,0.1,0.2,0.25\}$ within three stratified folds of the 296 training patients (15 fits), selecting $q=0.25$ by mean best balanced BCE. Five model types (support, continuous, zero, shuffled continuous, and real directed) were then fitted with mean and relational readouts and three seeds (30 fits). Training used 30 epochs, AdamW with learning rate 0.001 and weight decay 0.0001, batch size 16, gradient clipping at 5, and common patient-indexed schedules retaining one to four views. Validation BCE averaged equally over retained-view counts selected the earliest minimum checkpoint. No new model in this extension was evaluated on the old test partition.

All 15 nonempty view subsets of the 100 validation patients were scored. The development measure $R$ is AUROC averaged first across subsets of the same size and then equally across one-, two-, and three-view conditions. Probabilities from three seeds are averaged before calculating this measure. Missing-view labels remain the original patient-level labels. The same validation patients were used for checkpoint selection and evaluation; these are selected development estimates, not independent generalization estimates. The older benchmark had already examined the original test set and is retained as exploratory historical evidence, not repurposed as a new confirmatory test.

\subsection{What the observations support}
\begin{table}[htbp]
\centering
\caption{Preserved DeepDRiD development results on 100 validation patients used for checkpoint selection. Each estimate uses a three-seed probability ensemble. These are not independent test results.}
\label{tab:v07-real}
\begin{tabular}{lrrrr}
\toprule
& \multicolumn{2}{c}{$R$} & \multicolumn{2}{c}{Full-view AUROC}\\
Model & Mean & Relational & Mean & Relational\\
\midrule
Support phase & 0.9351 & 0.9367 & 0.9504 & 0.9548 \\
Continuous phase & 0.9403 & 0.9399 & 0.9568 & 0.9540 \\
Zero phase & 0.9367 & 0.9367 & 0.9552 & 0.9508 \\
Shuffled phase & 0.9387 & 0.9356 & 0.9512 & 0.9532 \\
Real directed & 0.9356 & 0.9342 & 0.9540 & 0.9488 \\
\bottomrule
\end{tabular}
\end{table}

With mean readout, continuous-phase $R$ was 0.9403 versus 0.9367 for zero phase and 0.9356 for real directed aggregation. The continuous-minus-zero difference was $+0.0036$, with a descriptive 95\% patient-bootstrap interval from $-0.0014$ to $+0.0095$; the continuous-minus-directed difference was $+0.0047$ ($-0.0030$ to $+0.0127$). Both include zero. Continuous minus shuffled phase was $+0.0016$ ($-0.0044$ to $+0.0072$). Continuous minus support phase was $+0.0052$ ($+0.0006$ to $+0.0110$), but the interval is conditional on selected models and is unadjusted for selection and multiplicity. It does not establish independent superiority.

The relational readout did not strengthen the main attribution: continuous-minus-zero $R$ was $+0.0032$ ($-0.0013$ to $+0.0093$), and the phase-by-readout interaction relative to zero phase was $-0.0004$ ($-0.0067$ to $+0.0068$). The intervals use 2,000 class-stratified patient bootstrap draws shared across models and subsets; repeated subset predictions do not increase the number of independent patients.

Continuous phase was mechanically active. Full-view mean-readout checkpoints had mean $\|\Im L\|_F/\|L\|_F=0.3494$ across seeds, with finite nonzero phase-only query gradients. This denominator differs from the normalized amplitude norm in \cref{eq:v07-normalized}. These diagnostics show an available phase pathway, not that its direction is clinically meaningful. Independent replay previously reproduced all 45 selected checkpoints and all 45,000 final validation probabilities.

\begin{figure}[htbp]
\centering
\includegraphics[width=\linewidth]{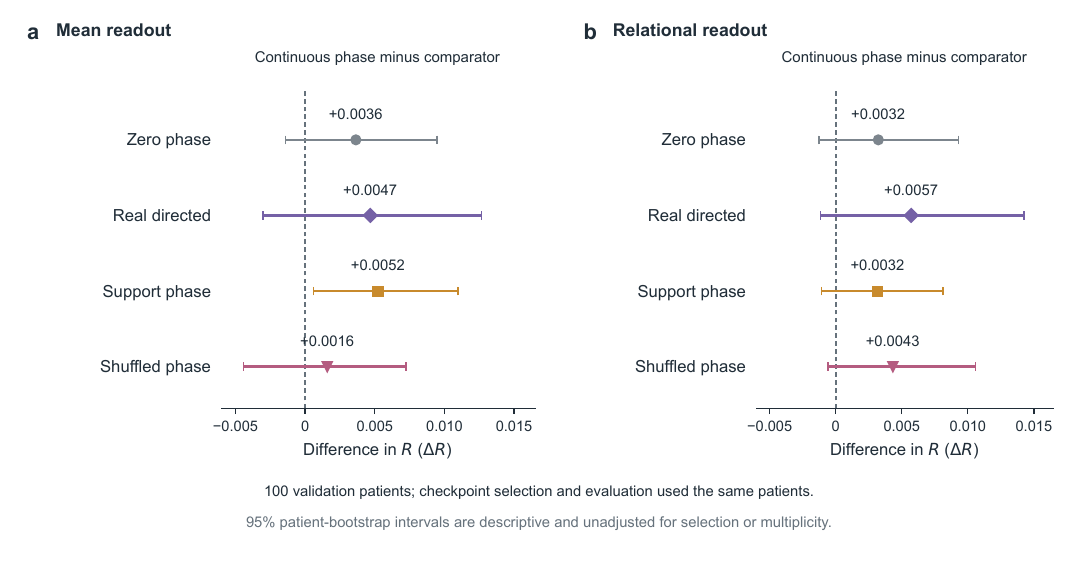}
\caption{Preserved DeepDRiD development contrasts. \textbf{a,b,} Continuous-phase minus comparator differences under mean and relational readouts, respectively. Colors and marker shapes identify the comparator consistently; both panels use the same horizontal scale. Points are three-seed ensemble estimates, with descriptive 95\% paired patient-bootstrap intervals; the dashed line is zero difference. All eight contrasts are shown. Checkpoint selection and evaluation used the same 100 validation patients, and intervals are unadjusted for selection or multiplicity. These selected development results do not establish independent clinical superiority.}
\label{fig:v07-real}
\end{figure}

DeepDRiD therefore supplies the third part of the argument: removing structural collapse is insufficient to establish task utility. The observations remain compatible with weak direction, an ineffective learned relation builder, readout/optimization limitations, limited sample size, or no material gain for this endpoint. They neither prove absence of direction in static images nor furnish a counterexample to every magnetic architecture.

\section{Discussion}
\subsection{Meaning of the claim}
The practical contribution is a sequence of questions with different evidential requirements. First, can the constructed operator distinguish the relevant orientations? The exact collapse identities answer a structural part of this question without training. Second, can the implementation use a directional signal when one is supplied? Paired opposite-label examples provide an identifiable positive control and expose chance-level limits of the erased representation. Third, does the learned direction improve the target application? That requires controlled comparisons on appropriate independent data. These questions should not be answered by a single end-to-end AUROC.

The advantage established here is specific: phase can preserve orientation that a symmetrized representation discards, and score-based phase can avoid the loss caused by reciprocal binary support. This supplies a reason to use magnetic encoding on a suitable task. It does not supply a reason to prefer it to every real-valued directed model. The directed comparator tests that distinction explicitly. Nor is complex arithmetic alone responsible: the zero-phase control retains complex filters, separating operator direction from nominal complex capacity.

\subsection{Design implications}
A fixed neighbor budget should be audited against the distribution of observed node counts, especially under missing modalities or views. Reporting only the configured $k$ hides complete-support cases with $m\leq k+1$. Reporting $r_B$ alone can also be misleading; $r_E$ and $\|L_q-L_0\|_F$ quantify the actual retained phase energy. Charge endpoints require care, since $q=1/2$ has a zero imaginary part without necessarily equaling $L_0$. These diagnostics can identify a nonfunctional ablation before expensive training.

Continuous scores provide one way to preserve reciprocal directional differences. Their success depends on what scores encode. On our paired task scores reveal the ground-truth orientation; on the retinal task they are learned from images without directional supervision. Bridging that gap requires tasks with independently justified directional relations, such as an observed ordering or a measured flow, followed by evaluation that does not reuse selection data. A nonzero phase on an arbitrary similarity graph is not itself a meaningful scientific mechanism.

\subsection{Limits and remaining evidence}
The exact identities follow directly from the chosen operator and should be judged as useful diagnostics, not a broad new theory of all magnetic GNNs. The synthetic benchmark has one small ring family, supplied edge scores, a local propagation label, fixed charges, and a simple direction-free control. Perfect separation is possible by construction. It validates the stated mechanism rather than realistic noise tolerance, multi-hop expressivity, large-graph scalability, or learned direction recovery. Seed ranges do not estimate population uncertainty. The models are controlled implementations rather than exhaustive tuned competitors; effective capacity differs between real and complex variants.

The clinical component is small, retrospective, image-only, and subject to development selection and prior test access. It has no untouched external cohort and supplies no clinical superiority claim. Moreover, $L_q\ne L_0$ is only an operator distinction. A particular input or readout can ignore it, and nonzero local phases need not imply nontrivial cycle flux or a distinct spectrum under every gauge transformation. We do not infer utility from spectral change or claim full-network gauge invariance.

\section{Conclusion}
Reciprocity can suppress a magnetic model's intended directional mechanism before learning. Exact energy identities and the top-$k$/observed-node boundary make that suppression measurable. Paired propagation controls establish where retaining direction can help, while the DeepDRiD comparison shows why structural availability must be followed by application-specific validation. The central claim is therefore a testable boundary of representation and utility, with a concrete remedy for score information lost by reciprocal support.

\section*{Declarations}
\textbf{Data and code availability.} DeepDRiD is available from its authors' public repository \citep{liu2022deepdrid}. The project retains the source revision and license, download/audit tools, image and manifest fingerprints, frozen-feature provenance, all configurations, selected checkpoints and histories, patient-level predictions, and scripts for the reported figures and tables. Data retain the source CC BY-SA 4.0 license. A permanent public code archive and DOI have not yet been assigned.

\textbf{Ethics and consent.} This is a secondary analysis of the released public dataset with no new recruitment or clinical intervention. Original collection and consent arrangements are described by the dataset authors. No new institutional approval or exemption is claimed; the submitting institution must determine and document applicable requirements.

\textbf{Authorship, funding, and competing interests.} Dawei Han, Zicheng Wang, and Yaoxing Ren contributed equally. Rongqian Wu and Fumin Wang are corresponding authors. Detailed author contributions, funding, and competing-interest declarations require confirmation by the responsible investigators before submission. No funding award or absence of competing interests is asserted on their behalf.

\textbf{AI assistance.} AI tools assisted with implementation, analysis scripting, numerical checks, and manuscript drafting. The responsible authors must review the evidence, interpretation, and applicable journal disclosure requirements. This research draft is not a clinical decision tool or a finalized submission.

\bibliographystyle{unsrtnat}
\begingroup
\interlinepenalty=10000
\bibliography{references}
\endgroup
\clearpage
\appendix
\section{Reproducibility and complete results}
The synthetic protocol is stored in \path{configs/direction_v07.json}; the generator and model are in \path{src/magnn_retina/direction_benchmark.py}. The training and independent replay/report entry points are \path{analysis/train_direction.py} and \path{analysis/report_direction.py}. Results include generated tensor splits, source/dependency identity, all selected checkpoints and epoch histories, per-graph test probabilities, all 72 run summaries, independent verification, and vector figures. Dataset generation seeds for train, validation, and test are 2026092101, 2026092102, and 2026092103. Initialization seeds are 20260921, 20260922, and 20260923. The complete test grid is shown below; no cell or seed has been excluded.

\begingroup\small
\setlength{\tabcolsep}{3.5pt}
\begin{longtable}{lllr r r r r r}
\caption{Complete synthetic test results. A21--A23 are AUROC and C21--C23 are accuracy for seeds 20260921--20260923. AUROC is shown to five decimals to expose the small finite-precision deviation described in the main text.}\\
\toprule
Task & $k$ & Model & A21 & A22 & A23 & C21 & C22 & C23\\
\midrule\endfirsthead
\toprule
Task & $k$ & Model & A21 & A22 & A23 & C21 & C22 & C23\\
\midrule\endhead
Direction & 1 & Zero phase & 0.50000 & 0.50000 & 0.50000 & 0.5000 & 0.5000 & 0.5000 \\
Direction & 1 & Support phase & 1.00000 & 1.00000 & 1.00000 & 1.0000 & 1.0000 & 1.0000 \\
Direction & 1 & Continuous phase & 1.00000 & 1.00000 & 1.00000 & 1.0000 & 1.0000 & 1.0000 \\
Direction & 1 & Real directed & 1.00000 & 1.00000 & 1.00000 & 1.0000 & 1.0000 & 1.0000 \\
Direction & 2 & Zero phase & 0.50000 & 0.50000 & 0.50000 & 0.5000 & 0.5000 & 0.5000 \\
Direction & 2 & Support phase & 0.50000 & 0.50000 & 0.50000 & 0.5000 & 0.5000 & 0.5000 \\
Direction & 2 & Continuous phase & 1.00000 & 1.00000 & 1.00000 & 1.0000 & 1.0000 & 1.0000 \\
Direction & 2 & Real directed & 1.00000 & 1.00000 & 1.00000 & 1.0000 & 1.0000 & 1.0000 \\
Direction & 7 & Zero phase & 0.50000 & 0.50003 & 0.50000 & 0.5000 & 0.5000 & 0.5000 \\
Direction & 7 & Support phase & 0.50000 & 0.50003 & 0.50000 & 0.5000 & 0.5000 & 0.5000 \\
Direction & 7 & Continuous phase & 1.00000 & 1.00000 & 1.00000 & 1.0000 & 1.0000 & 1.0000 \\
Direction & 7 & Real directed & 1.00000 & 1.00000 & 1.00000 & 1.0000 & 1.0000 & 1.0000 \\
Feature & 1 & Zero phase & 1.00000 & 1.00000 & 1.00000 & 1.0000 & 1.0000 & 1.0000 \\
Feature & 1 & Support phase & 1.00000 & 1.00000 & 1.00000 & 1.0000 & 1.0000 & 1.0000 \\
Feature & 1 & Continuous phase & 1.00000 & 1.00000 & 1.00000 & 1.0000 & 1.0000 & 1.0000 \\
Feature & 1 & Real directed & 1.00000 & 1.00000 & 1.00000 & 1.0000 & 1.0000 & 1.0000 \\
Feature & 2 & Zero phase & 1.00000 & 1.00000 & 1.00000 & 1.0000 & 1.0000 & 1.0000 \\
Feature & 2 & Support phase & 1.00000 & 1.00000 & 1.00000 & 1.0000 & 1.0000 & 1.0000 \\
Feature & 2 & Continuous phase & 1.00000 & 1.00000 & 1.00000 & 1.0000 & 1.0000 & 1.0000 \\
Feature & 2 & Real directed & 1.00000 & 1.00000 & 1.00000 & 1.0000 & 1.0000 & 1.0000 \\
Feature & 7 & Zero phase & 1.00000 & 1.00000 & 1.00000 & 1.0000 & 1.0000 & 1.0000 \\
Feature & 7 & Support phase & 1.00000 & 1.00000 & 1.00000 & 1.0000 & 1.0000 & 1.0000 \\
Feature & 7 & Continuous phase & 1.00000 & 1.00000 & 1.00000 & 1.0000 & 1.0000 & 1.0000 \\
Feature & 7 & Real directed & 1.00000 & 1.00000 & 1.00000 & 1.0000 & 1.0000 & 1.0000 \\
\bottomrule
\end{longtable}
\endgroup

\section{Additional real-data implementation details}
The retrospective retinal fits remain those of version 0.6; version 0.7 adds no patient observations or new retinal fitting. The charge screen used seed 20260920 in each of three training-patient folds. Its mean best balanced BCE values were 0.3827, 0.3771, 0.3777, 0.3755, and 0.3748 for ascending charges; these are tuning statistics. Final training seeds were 20260920--20260922. The shuffled-phase control permutes upper-triangle phase values within the observed edge union, restores antisymmetry, and uses a deterministic patient/seed permutation without label information. The control is retrained, so independently learned amplitudes can differ.

The relational readout concatenates mean real features, mean imaginary features, mean magnitude, and a per-channel phase-aligned edge correlation. For $u_{ic}=z_{ic}/\max(|z_{ic}|,10^{-8})$ and $\widetilde S=S/\sum_{ij}S_{ij}$, the last statistic is $\sum_{ij}\widetilde S_{ij}\Re\{\overline u_{ic}e^{\mathrm{i}\Theta_{ij}}u_{jc}\}$. Isolated-graph edge statistics are zero. The real directed control uses zero imaginary features and zero phase for this statistic. Magnetic models have 428,673 trainable parameters and real directed models 430,721, excluding the common frozen backbone. Equal nominal readout layout does not imply equal active head columns.

The source license and revision, quarantined-patient audit, patient partitions, frozen-feature provenance, original v0.5 and v0.6 reports, and prior prediction files are retained. The 100 validation patients comprise 50 negative and 50 positive cases. Quarantined training patient IDs were 56, 77, 164, and 167. Their removal was based on metadata ambiguity rather than outcomes or predictive performance. Source repository labels and licensing are not substitutes for institution-specific submission and ethics review.

\section{Interpreting related but different diagnostics}
For fixed $0<q<1/2$ and positive selected weights, three conditions are equivalent for support phase: reciprocal support, zero imaginary operator, and equality to $L_0$. At $q=1/2$ only the first and third remain equivalent. At $q=0$ all graphs have equality to $L_0$ regardless of support. At empty support $E_A=0$, both operators are zero but normalized reciprocity and distance ratios are undefined. These cases should be handled explicitly rather than assigned a misleading ratio.

For a continuous phase, reciprocity of binary support is not a collapse criterion. The relevant quantities are the retained antisymmetric score differences and the weighted sine terms in \cref{eq:v07-cont-energy}. For either construction, an operator norm does not quantify a trained predictor's accuracy, and a small norm alone gives no bound on its output without assumptions about filter and readout norms. A missing-view experiment changes both the graph and available visual information; it cannot isolate a causal clinical mechanism.

\end{document}